\documentclass[
    conf,
    table, xcdraw 
]{new-aiaa}
\usepackage[utf8]{inputenc}

\usepackage{graphicx}
\usepackage{amsmath}
\usepackage[version=4]{mhchem}
\usepackage{siunitx}
\usepackage{longtable,tabularx}
\usepackage{multirow}
\usepackage{placeins}
\usepackage[acronym,nonumberlist]{glossaries}
\usepackage{soul}
\usepackage{threeparttable}
\usepackage{booktabs}
\usepackage{colortbl}
\usepackage{gensymb}

\title{Air Traffic Controller Task Demand via Graph Neural Networks: An Interpretable Approach to Airspace Complexity}

\author{Edward Henderson\footnote{Research Associate, Project Bluebird, The Alan Turing Institute.}, Dewi Gould* and Richard Everson\footnote{Principal Investigator, Project Bluebird, The Alan Turing Institute.}}
\affil{The Alan Turing Institute, London, England NW1 2DB, United Kingdom}

\author{George {De Ath}\footnote{Lecturer, Department of Computer Science, University of Exeter.}}
\affil{University of Exeter, Exeter, England EX4 4QJ, United Kingdom}

\author{Nick Pepper\footnote{Senior Research Associate, Project Bluebird, The Alan Turing Institute.}}
\affil{The Alan Turing Institute, London, England NW1 2DB, United Kingdom}

\begin{document}

\maketitle

\begin{abstract}
Real-time assessment of near-term Air Traffic Controller (ATCO) task demand is a critical challenge in an increasingly crowded airspace, as existing complexity metrics often fail to capture nuanced operational drivers beyond simple aircraft counts. This work introduces an interpretable Graph Neural Network (GNN) framework to address this gap. Our attention-based model predicts the number of upcoming clearances, the instructions issued to aircraft by ATCOs, from interactions within static traffic scenarios. Crucially, we derive an interpretable, per-aircraft task demand score by systematically ablating aircraft and measuring the impact on the model's predictions. Our framework significantly outperforms an ATCO-inspired heuristic and is a more reliable estimator of scenario complexity than established baselines. The resulting tool can attribute task demand to specific aircraft, offering a new way to analyse and understand the drivers of complexity for applications in controller training and airspace redesign.
\end{abstract}

\section*{Nomenclature}

{\renewcommand\arraystretch{1.0}
\noindent\begin{longtable*}{@{}l @{\quad=\quad} l@{}}
ATC  & Air Traffic Control \\
ATCO  & Air Traffic Control Officer\\
FL  & Flight Level \\
CFL & Cleared Flight Level\\
XFL & Exit Flight Level\\
GAT  & Graph Attention Layer \\
GCS & Global Co-ordinate System \\
GNN  & Graph Neural Network \\
LMS  & London Middle Sector \\
MAE  & Mean Absolute Error \\
ML  & Machine Learning \\
TLPD  & Traffic Load Prediction Device \\
XGBoost  & Extreme Gradient Boosting \\

\end{longtable*}}

\vspace{20mm}
\noindent Author Accepted Manuscript version of paper at the AIAA AVIATION Forum 2025.

\newpage 

\section{Introduction}

\lettrine[loversize=0.1]{A}{ir} Traffic Control (ATC) ensures the safe, orderly and efficient flow of aircraft through controlled airspace by issuing instructions, or \textit{clearances}, to pilots based on radar and scheduling information~\cite{icao2001doc, Mats_part1}. Air Traffic Control Officers (ATCOs) are responsible for managing aircraft within designated three-dimensional regions of airspace (sectors), and must maintain safety standards while enabling timely progression along routes in the filed flight plan. Increasing traffic volume~\cite{easr_2022} requires greater automation and decision support to help manage ATCO task demand and maintain safety~\cite{SESARAutomationATM, Metzger2005, Antulov-Fantulin_Juričić_Radišić_Çetek_2020}. A key challenge lies in helping ATCOs prioritise attention and issue clearances effectively, particularly during periods of high traffic and communication constraints. A robust and accurate measure of upcoming controller burden is an essential prerequisite for this challenge. In this work, we leverage key insights from ATCOs to construct a graph-based surrogate measure for short-term task demand. In particular, we train a Graph Neural Network (GNN)~\cite{gnn} to predict upcoming clearances and carry out a systematic ablation study in order to define a flexible, robust, and nuanced measure of near-term operational demands.

A variety of approaches have been proposed to quantify the complexity of ATC. For instance, motion-based metrics have been introduced in works such as~\cite{5641823, Wee_Lye_Pinheiro_2018}, where complexity is derived from aircraft trajectory vectors. A probabilistic treatment of complexity, particularly focused on weather-dependent events, is presented in~\cite{aerospace11100798}. Alternatively, some approaches model the airspace, and the aircraft within, as a graph, allowing topological metrics like strength, clustering coefficient, and nearest neighbour degree to serve as proxies for complexity~\cite{Isufaj2021}.

Recent work has also explored data-driven methods. In~\cite{aerospace9120758, CHEN2025110090} deep learning models are trained on image-based representations of air traffic to learn complexity features. Complementary to these efforts, several studies have focused on general-purpose decision support tools for air traffic management~\cite{10749617, aerospace8090260}, while others have targeted more specific applications. Examples of this include sector-specific conflict resolution~\cite{guleria2021machine}, and broader autonomous systems that replicate significant components of tactical ATC operations~\cite{brittain2019autonomous, brittain2022scalable}.

In this work, we provide a new perspective on short-term controller task demand and airspace complexity, centred around the concept of \textit{upcoming clearances}. We hypothesise that the set of clearances that will need to be issued in the near future provides a reliable and operationally meaningful surrogate for controller burden. ATCOs excel at rapidly identifying which aircraft require immediate attention, based on both local conflicts and broader contextual awareness. Replicating this decision-making and prioritisation in machine learning models is challenging, particularly when the underlying interactions between aircraft can be highly non-trivial. 

A graph-based representation offers a natural and interpretable structure for modelling these interactions, allowing the model to reason not only about individual aircraft but also about the relationships between them. Furthermore, this representation facilitates the incorporation of qualitative insights from ATCO interviews, enabling the integration of domain knowledge into the model design. While not intended for operational deployment, the method serves as a research tool to better understand airspace complexity and controller task demand. It is especially well-suited for use in ATCO training, providing insight into sector dynamics in both real-world and synthetic environments, and potentially supporting airspace redesign efforts.

We first train a GNN to predict the future number of clearances an ATCO will issue in a static air traffic scenario. We show that our method significantly outperforms an ATC heuristic baseline and two popular regression models (Random Forest~\cite{Breiman2001} and an Extreme Gradient Boosting (XGBoost)~\cite{Chen2016}). 
These results demonstrate and validate our use of a graph-based representation and architecture.

While predicting upcoming clearances is a crucial validation step, it serves as a proxy for our ultimate objective: modelling ATCO task demand. The number of clearances, while strongly correlated with task demand, is an incomplete measure of complexity. As an example, there are potentially multiple valid clearance sequences that result in safe and efficient outcomes, and a complex air traffic interaction may be solved with few, but well-designed clearances from the ATCO. Additionally, an aircraft receiving clearances may not be the source of complexity in a situation. Consequently, using the number of clearances as a stand-alone metric for controller burden introduces noise and potentially obscures the full picture of controller decision-making. Motivated by this, we introduce a surrogate measure of controller task demand derived from a systematic ablation methodology: we quantify the change in the number of predicted clearances as individual aircraft are computationally removed from a scenario (see, e.g.,~\cite{8967743, roelofs2022causalagents} for related work in the autonomous vehicle literature).

A persistent challenge in the airspace complexity literature is the ambiguous mapping between analytical indicators and the actual task demand experienced by ATCOs. Current methods often rely on direct ATCO labelling for model optimisation~\cite{CHEN2025110090} or fall back on simplistic proxy metrics, such as aircraft count, for baselines~\cite{Isufaj2021}. Here, we validate our task-demand measure \textit{qualitatively}, showcasing its effectiveness across a variety of real-world scenarios and contrasting its performance with these established metrics.

A key benefit of our approach is its \textit{interpretability}. Our model not only provides a measure of future controller task-demand but, using systematic ablation, can also attribute that demand to individual aircraft within an air traffic scenario. This attribution capability enhances the tool's practical usability while offering a robust method for validating its results.\looseness=-1

In summary, the key contributions of this work are threefold:

\begin{enumerate}
\item \textbf{Graph-based Scenario Representation.} We propose a novel graph-based representation of air traffic that encodes both individual aircraft features and their influence upon one another. This structure provides a rich, contextual model of the spatial and operational relationships between aircraft.

\item \textbf{Learning to Predict Upcoming Clearances.} We develop a Graph Neural Network (GNN) trained on this representation to predict the number of upcoming clearances for each aircraft. This approach significantly outperforms a heuristic method developed following ATCO interviews and demonstrates competitive performance against established regression baselines.

\item \textbf{An Interpretable Surrogate Measure of Controller Task Demand.} We define a principled, data-driven surrogate measure for controller task demand by applying systematic ablation to the trained GNN. This measure captures both traffic complexity and controller action anticipation. We then demonstrate its superiority to existing metrics by presenting its performance in real-world air traffic scenarios.
\end{enumerate}

\section{Methodology}
In this section, we first describe the data used for this study and explain how we construct the scenario graph representations with detailed examples. Secondly, we discuss the model architecture and implementation used to train the GNN to predict upcoming clearances. We describe the feature importance study we performed on our model, as well as how we quantitatively compared our method to two benchmark methods. Lastly, we discuss how we construct our surrogate measure of ATCO task demand and how this method was compared against existing approaches.

\subsection{Data}
\label{ssec:data}
For this study, we collected data from July to September 2019 within the London flight information region~\cite{nats_fir}. The dataset comprised radar tracks from secondary radar which is a transponder-based call-and-response radar with extra information fields in addition to simple primary positional information. Additionally, scheduled flight data, ATCO-issued clearances, and details of the coordination agreements that were made between controllers of neighbouring sectors were included in the dataset. This study focuses on a specific sector in the UK's en-route airspace, London Middle Sector (LMS), with an altitude range between flight levels (FLs\footnote{Flight levels are expressed as multiples of 100 feet.}) 215 and 305. The dataset was filtered to include only those aircraft that passed through LMS, resulting in a dataset of 46{\small,}570 trajectories and 116{\small,}614 associated clearances issued to those aircraft.

\begin{figure}[ht]
\centering
\includegraphics[width=0.96\textwidth]{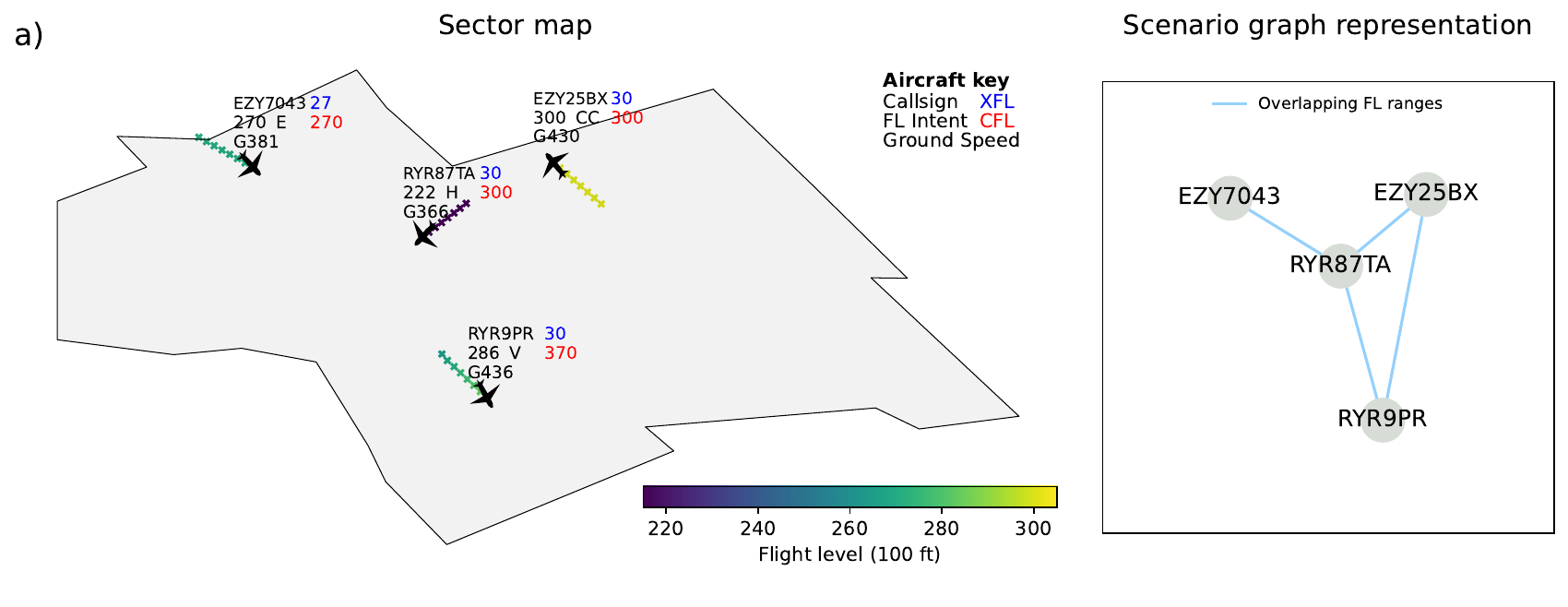}\\
\vspace{-2mm}
\includegraphics[width=0.96\textwidth]{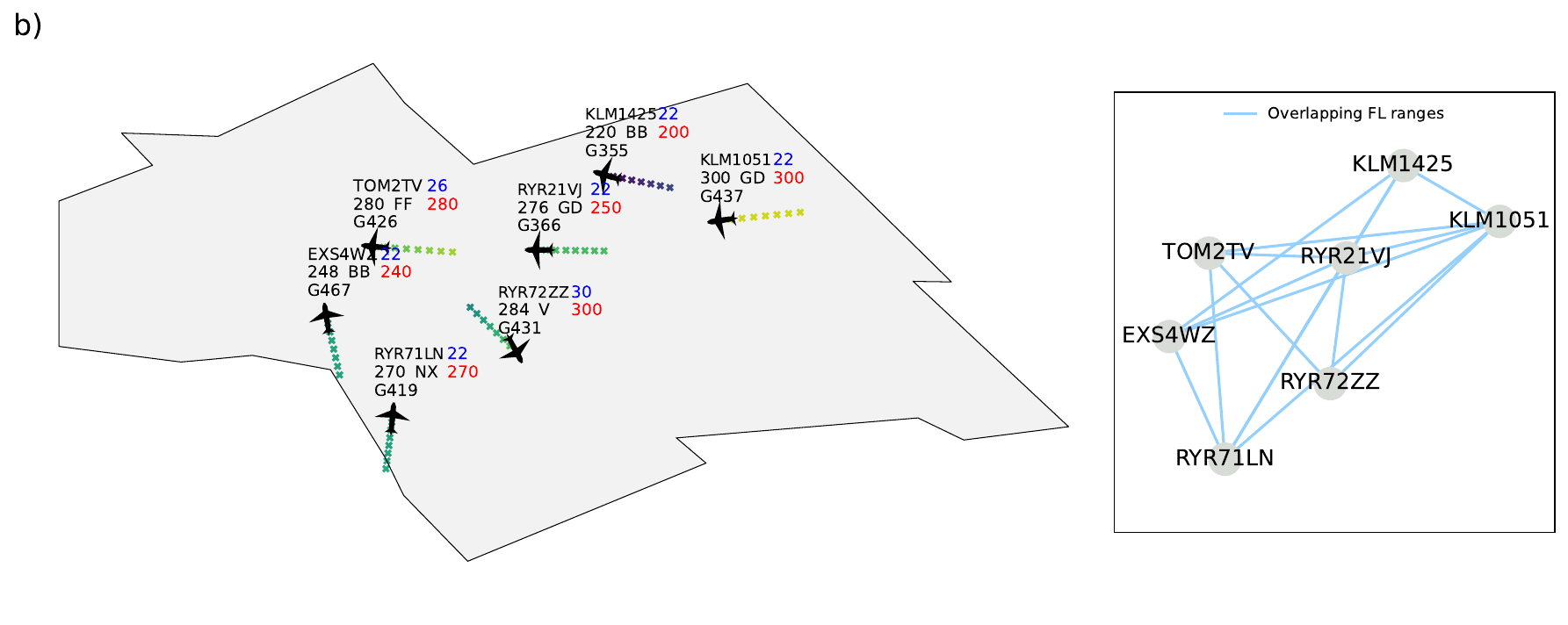}
\caption{Two example scenarios in July 2019 from the historic LMS dataset alongside the graph-based representations. In the left-hand plots, the black outlined shape is the boundary of the LMS airspace, aircraft are indicated with icons and their past radar tracks shown with the colour corresponding to altitude. Corresponding graph representations of the scenarios are shown in the right-hand plots. The graph nodes, which represent aircraft, are connected by edges if there is an overlap in the ranges of their current and potential future flight levels. The traffic level in example a) is relatively quiet, whereas the sector is busier in example b).}\label{fig:scenarios}
\end{figure}

The chosen sector is located above the London Terminal Manoeuvring Area, which includes departures at and arrivals from several of the world's busiest airports. The majority of aircraft within LMS are either climbing or descending. On the left-hand side of Fig.~\ref{fig:scenarios} we present the sector geometry with two example scenarios. Generally speaking, aircraft flying southwards in LMS are climbing to their cruising altitude, such as RYR9PR in Fig.~\ref{fig:scenarios}. Aircraft travelling to the west are usually flying to Bristol or Cardiff airports (TOM2TV and RYR21VJ). On the other hand, those travelling to the north of the sector, such as KLM1425 and RYR71LN, are usually descending towards regional airports in the UK Midlands.

In this work, we define a \textit{scenario} as a static snapshot of all the aircraft in LMS at a particular time, with aircraft position derived from the most recent radar sweep, following their most recent clearance issued by ATC. ATCOs are responsible only for the aircraft within their sector; however, they will also be aware of aircraft in adjacent sectors that are about to enter, as well as those they have just handled and have handed off to the next sector. Therefore, aircraft within a 0.3$\degree$ buffer region ($\sim$10-18NM) of the LMS sector boundary, that will or have already passed through LMS airspace, are also included in the scenario snapshots. Air traffic scenarios were captured at three-minute intervals throughout the three-month period. Scenarios were only extracted between the times of 05:30-18:30 each day, as the LMS sector was regularly combined (band-boxed) with another sector outside of these times and controlled by a single ATCO due to lower traffic. Using the above dataset, for a given scenario we had access to position, flight level, velocity, climb rate, aircraft type, flight plan and coordination data for all aircraft. Scenarios which contained no aircraft were excluded. This resulted in $\sim$21{\small,}000 scenario data samples for our analysis. Data from a week in August 2019 (11th - 17th) was reserved as a held out test set of 1{\small,}550 scenarios.

The goal of the model was to predict the number of clearances that would be issued to each aircraft in a scenario within a given time period. This prediction is for aircraft currently within the sector, excluding aircraft that enter within the time period, resulting in an estimation of instantaneous task demand for the current scenario. Recognising that there are a number of possible clearances in the standard phraseology used in en-route operations (see, e.g. CAP 413~\cite{CAP413}), we restrict the set of predicted clearances to those that directly impact an aircraft's trajectory within LMS. For instance, clearances concerning position reporting or related to co-ordination with neighbouring sectors are not included in this study. Table~\ref{tab:clearances} details the types of clearances used within the dataset. A forecast period of $10$ minutes was chosen for this initial research.

\newcolumntype{L}[1]{>{\raggedright\arraybackslash}p{#1}}  
\begin{table}[h!]
\centering
\footnotesize
\begin{threeparttable}
\caption{\label{tab:clearances} Types of clearances predicted by the task demand model}
\begin{tabularx}{\linewidth}{L{3.2cm} X}
\rowcolor[RGB]{239,239,239}
\toprule
\textit{Clearance type} & \textit{Description} \\
\midrule
 Heading instructions & Instruction to fly on either a specified heading or to redirect towards a fix. \\
 Level instructions & Instruction to climb or descend to a specified flight level either  \textit{immediately} or at the pilot's discretion. The ATCO may also specify that the climb or descent be achieved by a specified location. \\
 Speed instructions & Instruction to fly above or below a specified speed. \\
\bottomrule
\end{tabularx}
\end{threeparttable}
\end{table}

\subsubsection{Graphical representation of air traffic scenarios}
\label{sssec:graph_features}
Figure~\ref{fig:scenarios} shows examples of the graph representation for two scenarios in LMS. Each air traffic scenario was processed and represented as an undirected graph $G = (V,E)$ with a set of nodes, otherwise known as vertices, $V$ and a set of undirected edges $E$ joining pairs of nodes. Nodes pertain to individual aircraft within the scenario. Each node $v_i \in V$ has an associated ${d_V}$-dimensional feature vector $\mathbf{x}_i \in \mathbb{R}^{d_V}$ containing data attributes for that aircraft.

\begin{figure}[ht]
\centering
\includegraphics[width=0.67\textwidth]{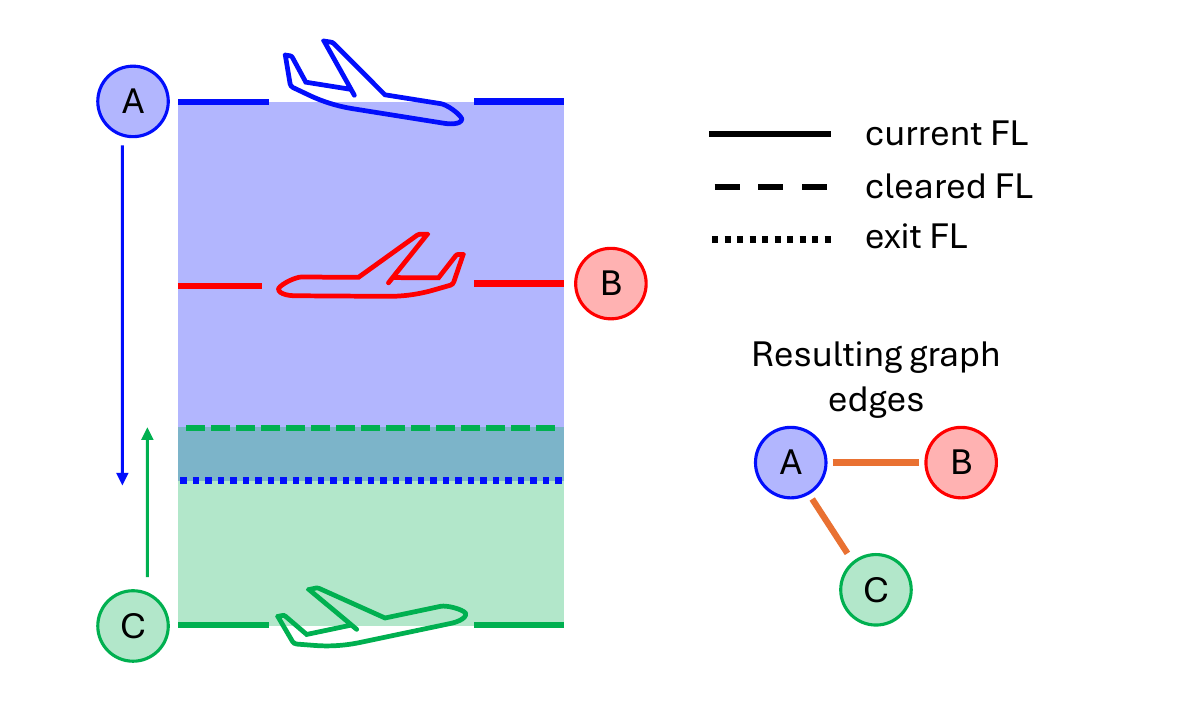}
\caption{Example of the edge creation process. Coloured regions represent the vertical flight level ranges of each aircraft. The flight level ranges of both A and B, and A and C overlap, resulting in these nodes being connected with edges in the graph representation. Notice that the ranges of B and C do not overlap, and, therefore, the corresponding nodes are not connected with an edge.}\label{fig:overlapping_FLs}
\end{figure}

The edges in our graph-based representation are interaction terms between aircraft within the scenario. Edges are created between aircraft nodes that are deemed to have potential interactions from an ATCO decision-making perspective. Specifically, if the vertical flight level ranges of two aircraft overlap, an edge is created between the relevant nodes. To evaluate this, we look at the current flight level (FL), the current cleared flight level (CFL) which the aircraft may be ascending or descending toward, and the exit flight level (XFL) which must be reached before exiting the current airspace sector. The span of these three values corresponds to the vertical range that an aircraft may occupy. If the vertical ranges of two aircraft intersect, we consider them potential interactions and add an edge\footnote{A buffer of ten flight levels is included in this computation.}. Figure~\ref{fig:overlapping_FLs} shows a graphical example of the edge creation process.

Similar to the nodes, each edge $e_{ij} \in E$ has an associated feature vector $\mathbf{x}_{ij} \in \mathbb{R}^{d_E}$ containing data attributes pertaining to the interaction term, where $d_E$ is the dimensionality of the edge feature vectors (i.e., the number of data attributes).\looseness=-1

Table~\ref{tab:features} summarises the computed edge and node data attribute features. The selection of these features was guided by interviews with ATCOs.

\begin{table}[ht]
\centering
\footnotesize
\begin{threeparttable}
\caption{\label{tab:features} Descriptions of the data features encoded into the graph representations of scenarios.}
\begin{tabularx}{\linewidth}{L{3.2cm} L{1.4cm} X}
\rowcolor[RGB]{239,239,239}
\toprule
\textit{Feature} & \textit{Node / Edge} & \textit{Description} \\
\midrule
Lateral position & node & 2D horizontal position of the aircraft in Global Co-ordinate System (GCS) co-ordinates. \\
Flight level & node & Aircraft altitude in increments of 100 feet. \\
Ground speed & node & Horizontal speed of the aircraft across the ground in knots. \\ 
Lateral direction & node & 2D horizontal velocity components. \\
Climb rate & node & Rate of vertical ascent or descent in feet per minute. \\
Step climb & node & Boolean indicating if the aircraft had been climbing or descending in steps\tnote{a}. \\
Delta to exit FL & node & Difference between current and desired flight level at sector exit in increments of 100 feet. \\
Engine type & node & Aircraft engine type (piston, turboprop or jet). \\
Wake category & node & ICAO wake turbulence category (light, medium, or heavy/jumbo). \\
On heading & node & Boolean indicating if aircraft is on a heading instead of its filed route. \\
Speed control & node & Boolean indicating if a speed condition was issued. \\
Comm state & node & Boolean indicating whether the aircraft is in communication with sector controllers. \\
Time to exit & node & Expected time remaining in the current sector in seconds\tnote{b}. \\
Sector exit direction & node & Anticipated sector exit direction (N, S, E, or W) from the flight plan. \\ 
\midrule
Separation distance & edge & Distance between the aircraft in metres\tnote{c}.\\
Closing speed & edge & Component of relative velocity toward the aircraft pair's midpoint in knots. \\
\bottomrule
\end{tabularx}
\begin{tablenotes}
\footnotesize
\item[a] Aircraft were considered to be climbing or descending in steps if they had been issued multiple vertical clearances within the sector, or if their CFL differed from their desired XFL.
\item[b] Aircraft trajectories were simulated forward on their current heading or route using the BADA model~\cite{nuic2010bada} until sector exit.
\item[c] Euclidean distance calculated in Earth-Centred, Earth-Fixed coordinates~\cite{HofmannWellenhof2001}.
\end{tablenotes}
\end{threeparttable}
\end{table}

\subsection{Methods}
\subsubsection{Model architecture}
We used a graph neural network (GNN) to learn descriptive features about aircraft and their interactions in different scenarios. GNNs are a type of deep learning model that are specialised to process and learn relationships within graph-structured data by iteratively distributing and aggregating information across the graph using message-passing operations or layers~\cite{Zhou2020}. Information can be encoded within both the graph nodes and edges, making the approach very flexible. As such, a GNN-based approach was suitable to learn features from scenarios which can contain variable numbers of inputs, i.e., aircraft and interactions, at any one time. Two graph attention v2 (GATv2~\cite{Brody2022}) layers were used to leverage both the node features (aircraft characteristics) and edge features (interaction terms between the aircraft present in the sector). GATv2 layers apply an attention mechanism in order to assign varying importance to neighbouring nodes using both the graph structure and edge features.

\begin{figure}[h!]
\centering
\includegraphics[width=0.82\textwidth]{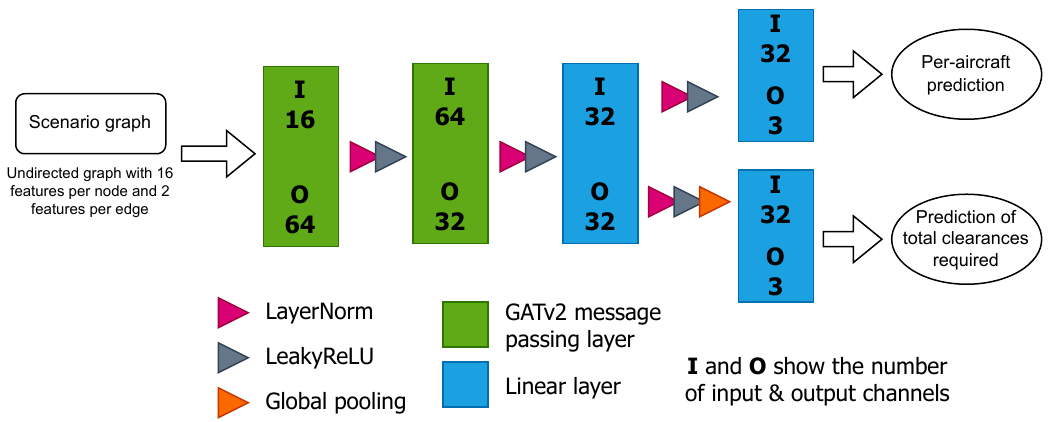}
\caption{Schematic showing the structure of the GNN prediction model. Two GATv2 layers are used to generate rich context-aware embeddings for each aircraft node using the graph representation of the input scenario. The model has two separate linear layer heads to produce total graph-level and per-aircraft predictions of the number of clearances issued within the ten-minute period following the input scenario.}
\label{fig:models}
\end{figure}

Edges between nodes in the scenario graph corresponded to potential interactions between aircraft. The GATv2 layer's attention mechanism assigns a dynamic weight to each edge, quantifying the relevance of each potential interaction. This allows the model to learn which of the initially defined connections represent the most operationally significant aircraft interactions. The context-aware node-wise embeddings, generated by the GATv2 layers, were passed through a series of linear layers to perform the desired regression predictions. Two prediction heads were used to generate both scenario-level (graph-level) and per-aircraft (node-level) predictions of the number of clearances that would be issued by an ATCO in the 10-minute period following the current input scenario. Jointly optimising for both graph- and node-level predictions provides fine-grained supervisory signals during training, for instance, which specific aircraft are associated with high clearance counts, while also enabling a clear interpretation of the final task demand attribution. This is expected to improve the quality of the learned node embeddings and scenario-level predictions on unseen data. The specific architecture of our proposed GNN model is shown in Fig.~\ref{fig:models}.

\subsubsection{Implementation details}
The models were implemented using \textit{PyTorch} 2.5.1~\cite{paszke2019pytorchimperativestylehighperformance} and \textit{PyG} 2.6.1~\cite{fey2019fastgraphrepresentationlearning}. Each of the GNN models were trained for a maximum of $50$ epochs, using a mini-batch size of eight and the AdamW optimiser (learning rate of 0.001)~\cite{Loshchilov2018}. Models were jointly optimised on both the per-aircraft and scenario-wide prediction tasks during the training phase. The model was trained to produce three predictions for each task, a predicted median number of clearances and estimates of the 10\textsuperscript{th} and 90\textsuperscript{th} percentile values. This quantile regression was performed using a pinball (quantile) loss function:
\begin{equation}
\mathcal{L}(y,\hat y)
  = \frac{1}{3}
    \sum_{q\in\{0.1,0.5,0.9\}}
      \max \Bigl(
        q (y - \hat{y}_q), \; (q - 1) (y - \hat{y}_q)
      \Bigr).
\end{equation}
where $\hat{y}_q$ are the models predicted values of the 10\textsuperscript{th} percentile, median and 90\textsuperscript{th} percentile values for each task~\cite{Koenker1978, chung2021pinballlossquantilemethods}.
The ground truth number of clearances is an inherently noisy supervision signal, as ATCOs can resolve a given scenario using multiple valid clearance sequences.  This makes it an unreliable target for direct mean value regression. Predicting multiple quantiles simultaneously provides a robust alternative by encouraging the model to satisfy multiple statistical constraints, e.g. $\hat{y}_{0.1} \leq \hat{y}_{0.5} \leq \hat{y}_{0.9}$. This acts as a form of implicit regularisation at training time, pushing the model to learn shape characteristics of the target distribution rather than precisely matching potentially noisy target values~\cite{Narayan2021}. Once trained, the median value prediction is the only desired output and is the prediction used for model evaluation.

Most numerical data features (e.g., flight level) were standardised using z-score normalisation, while ordinal, categorical and boolean features were encoded as integer classes. 
Features for which the zero value has an important physical meaning, the climb rate, delta to exit flight level, and closing speed, were scaled using maximum absolute scaling which preserves zero while mapping values to the range $[-1, 1]$, as:
\begin{equation}
x_{\text{scaled}} = \frac{x}{\max\left(|x_{\min}|, \; x_{\max}\right)}.
\end{equation}

\begin{figure}[h!]
\centering
\includegraphics[width=0.8\textwidth]{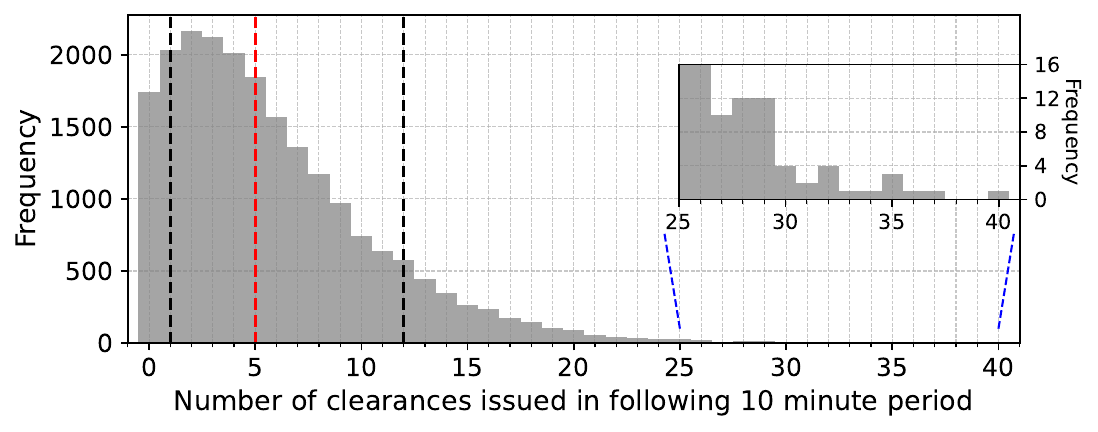}
\caption{Histogram showing the distribution of the number of clearances issued to aircraft in each of the scenario samples following the current air traffic snapshot. The 10\textsuperscript{th}, median and 90\textsuperscript{th} percentile values are indicated with black and red vertical dashed lines. An inset figure is included to show examples in the long upper tail of the distribution. It should be noted that this distribution shows the number of clearances issued to all aircraft present in the scenario at the beginning of the 10-minute period and excludes any clearances issued to aircraft that enter the sector during this time.}
\label{fig:clearances_distribution}
\end{figure}

The dataset was highly imbalanced with a long tail for scenarios with higher numbers of issued clearances, as shown in Fig.~\ref{fig:clearances_distribution}. Since these cases are potentially some of the most interesting and complex, a weighted sampling strategy was used when training the GNN model. Specifically, samples were uniformly drawn from four target value intervals of issued clearances: below the 10\textsuperscript{th} percentile, between the 10\textsuperscript{th} and 50\textsuperscript{th} percentiles, between the 50\textsuperscript{th} and 90\textsuperscript{th} percentiles and above the 90\textsuperscript{th} percentile. This approach meant the model was exposed equally to both rarer scenarios, where higher number of clearances were issued, and more common examples. As a result, both rare and frequent scenarios contribute more equally to the training loss, reducing the bias towards the majority of the distribution~\cite{Li2014, He2021}.

We trained the GNN using a five-fold cross-validation approach~\cite{Kuhn2013a}. Each of the five models had a dataset of 13{\small,}453 scenarios for training, 3{\small,}363 scenarios used for validation to determine optimal model weights during training and 4{\small,}205 scenarios for internal testing of that fold. By performing cross-validation, we could evaluate each trained GNN model on its respective test set, which allowed us to assess the stability of the learning task. The results of the cross-validation experiments are shown in the appendix. When finally evaluating our GNN approach on the reserved week of data (2{\small,}279 held out scenarios) we employed an aggregation ensemble approach to reduce prediction variance and improve performance~\cite{Kuhn2013b}.
The final predicted number of clearances required for scenarios in the reserved data was the mean of the five predictions made by the cross-validation models.

\subsubsection{Feature importance analysis}
\label{sec:featureimportantceanalysis}
To assess the relative importance of the node and edge features described in Section~\ref{sssec:graph_features}, a permutation analysis was conducted. The best performing of the five cross-validation models, as previously determined, was initially evaluated on its corresponding test dataset. Then, one-at-a-time, the values of each feature were randomly permuted across the test set of that cross-validation model. We then re-ran inference with the model and calculated the difference between the permuted and unperturbed dataset results. Each permutation test was repeated 20 times across the 4{\small,}205 testing scenarios, yielding 84{\small,}100 permuted scenarios. The mean absolute error (MAE) in the number of predicted clearances was computed across all permuted scenarios, and the change from the original result was computed. A feature is considered important if its permutation leads to a comparatively large reduction in model performance, shown by a marked increase in MAE. A decrease in MAE after permutation would imply that the feature was detrimental to the model's performance, providing noise rather than a useful signal.

To evaluate the importance of the graph structure and determine whether the outlined graph construction method was beneficial, we ran an ablation experiment with randomised graph connectivity for the scenario representations. Graph edges were assigned randomly using a binomial (Erdős-Rényi) graph generator~\cite{Erdos:1959:pmd}, with the probability of edge creation uniformly sampled from $[0, 1]$. Separation distance and closing speed edge features were computed for all randomly created edges. For every scenario in the datasets, 20 connectivity samples were generated. We additionally conducted a specific experiment in which all edges were removed from the graph scenario representations. The model performance when using randomly-connected and edgeless graphs was then compared with our original graph construction method.

\subsubsection{Quantitative comparison of prediction methods}
Multiple baseline methods were employed to provide reasonable comparisons to our proposed model. For each method, we computed the MAE between the predicted number of clearances issued to all aircraft in the test scenarios during the subsequent 10-minute period with the number actually issued by the responsible ATCO.

The first baseline is a simple, algorithmic method for predicting the number of clearances needed to resolve an air traffic scenario. In this baseline method, every aircraft in the scenario that is not already at its exit flight level for the sector, or cleared to ascend or descend toward it, contributes one clearance to the estimated total number. Such a method provides an estimation of the minimum number of clearances required to resolve an air traffic scenario by assuming there are no aircraft that would impede a single clearance to the exit flight level. We refer to this method as the \textit{minimum-clearance} baseline.

We additionally benchmarked our GNN approach against two standard regression methods, a random forest regressor~\cite{Breiman2001} and an Extreme Gradient Boosting (XGBoost) model~\cite{Chen2016}, to assess its performance level against established regression techniques. The same node features were used to train both regression models to predict the number of clearances required for each aircraft based solely on its own features. 

\subsubsection{Task demand estimation}
\label{sssec:taskdemandestimation}
Once trained, we used the GNN clearance prediction model to infer the influence of individual aircraft on overall ATCO task demand. To achieve this, we applied an ablation method that iteratively removed each aircraft from the scenario graph $G$ and re-computed the predicted total number of issued clearances. This provides a direct task demand score for each aircraft, $\phi_i$:
\begin{equation}
\label{eq:taskdemandeq}
    \phi_i = C(G) - C(G \setminus \{ v_i \}),
\end{equation}
where $C(G)$ is the predicted number of clearances given to scenario graph $G$, and $G \setminus \{ v_i \}$ is the scenario graph $G$ with aircraft node $v_i$ removed. This ablation-based score provides a more robust estimate of an aircraft's contribution to task demand than its per-aircraft clearance prediction alone. By systematically removing each aircraft, we can measure its marginal impact on the scenario's overall complexity.

\subsubsection{Qualitative task demand comparisons}
\label{sssec:quali_baselines}
Quantitatively assessing the ATCO task demand predictions of our GNN-based approach is challenging as it is difficult to establish a consistent ground truth for a subjective outcome. Moreover, similar methods for predicting and analysing airspace complexity cannot be directly compared as they measure different quantities. Therefore, we show, and qualitatively compare, estimated values of task demand from our GNN-based approach with two established approaches used for airspace complexity analysis across a four-hour span of operations in LMS. We now briefly outline the two approaches compared with our method.

The first approach was the Traffic Load Prediction Device (TLPD) developed by NATS Software Services~\cite{nats2010altran} and used as part of the tactical operational management systems in UK en-route ATC centres. The TLPD system shows aircraft traffic count predictions for individual airspace sectors up to four hours ahead of the current time. The TLPD also contains a linear complexity prediction model calibrated for every sector in UK area and terminal control. The complexity model included in TLPD is a weighted sum of several factors, some of which we include in our model, including the numbers of: aircraft in a sector; aircraft entering the sector; jet, prop jet and piston aircraft; aircraft climbing and descending; aircraft cruising in different flight level ranges; slow and high flights; and fast and low flights. Additionally, the TLPD models sector-specific hotspots with fixed additive values for flights using certain routes or travelling to specific locations or airports. The particular weights used in the linear model were calibrated for each sector and hotspot factors determined by sector experts. The predicted complexity value is intended to be compared to the TLPD traffic count to indicate a relative complexity for that volume of traffic. In practice, the TLPD traffic counts are shown as a histogram with the predicted complexity line overlaid on top (see Fig.~\ref{fig:qualitative_results} for an example). 

The task demand estimated by our GNN-based approach was also compared to the spatio-temporal graph indicators introduced by \citet{Isufaj2021}. Their work proposed metrics to analyse airspace complexity by modelling air traffic scenarios as graphs, with aircraft represented by vertices and edges created between aircraft that are within certain vertical and horizontal separation thresholds. These thresholds are determined by computing the mean pairwise horizontal (haversine) and vertical distances between aircraft from data for the previous day and adding buffer values. 
For our analysis in LMS, the separation thresholds dictating graph edge creation were calculated as 48 nautical miles horizontally and 4,400 feet vertically. Edges weights are scaled from zero at the minimum aircraft separation for edge creation to one at a standard safety distance of 5 nautical miles horizontally and 1,000 feet vertically. 
Four indicators were designed from common graph metrics: the \textit{edge density}, \textit{graph strength}, \textit{clustering coefficient} and \textit{nearest neighbour degree}. The authors asserted that analysing the evolution of these metrics over time offered useful insight into the complexity of the represented air traffic scenarios. 
The \textit{edge density} is a global connectivity measure, in this case, a single number for each scenario showing the proportion of aircraft that are close together. The strength is a per-vertex measure of centrality, corresponding to the number of aircraft within the scenario each is close to. The \textit{graph strength} indicator is calculated by taking the mean strength of all vertices. The \textit{clustering coefficient} measures how tightly connected the local neighbourhood of each vertex is, and its average value is used as an indicator of cluster structure. The \textit{nearest neighbour degree} is a weighted average of the degree of connected vertices. For complete details on their graph creation method and indicator definitions see~\cite{Isufaj2021}.

\section{Results}

\subsection{Feature importance quantification}
\begin{figure}[ht!]
\centering
\includegraphics[width=0.98\textwidth]{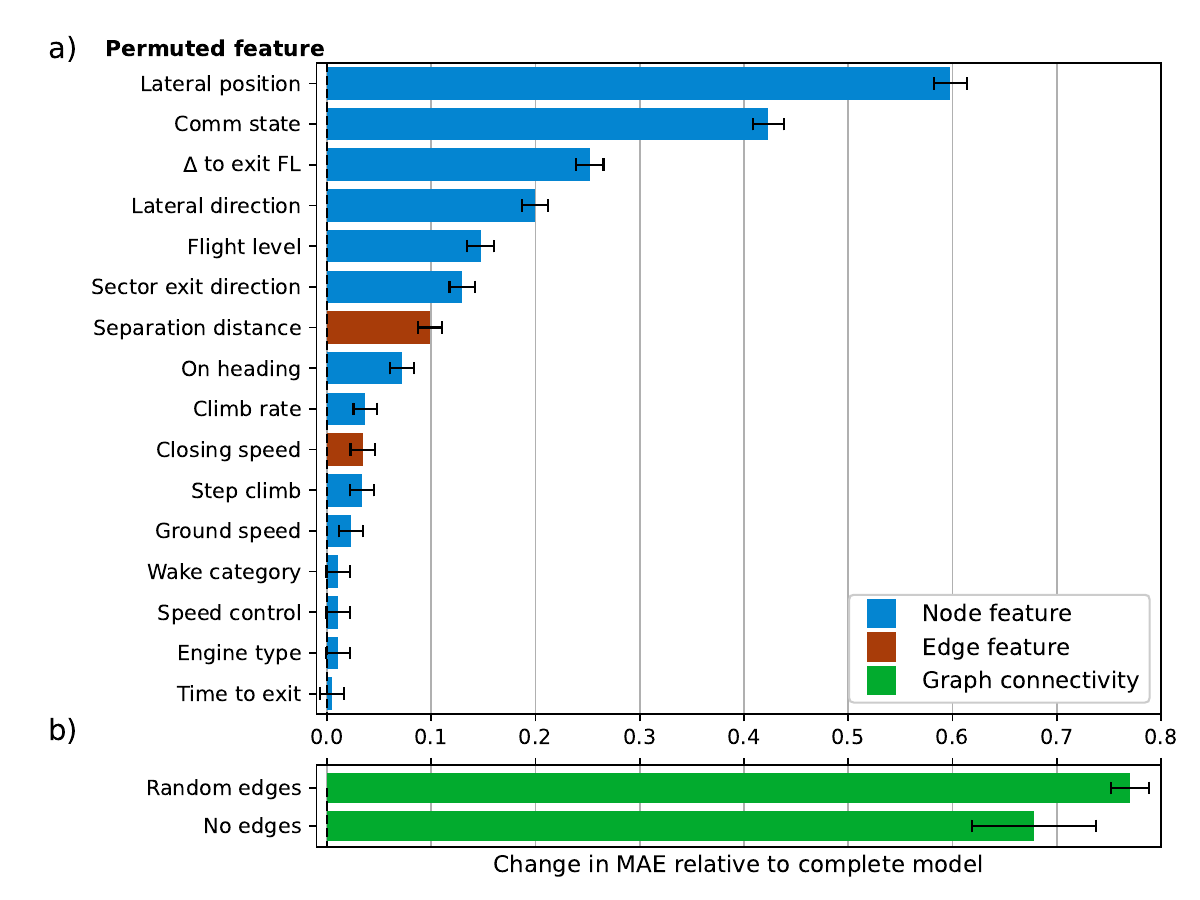}
\caption{a) Each horizontal bar shows the change in mean absolute error (MAE) when the value of a given feature is randomly permuted, compared to the model's performance on the unperturbed dataset. A large positive change in MAE indicates the feature is more important for our prediction model. A decrease in MAE after permutation would imply the feature was detrimental to the model's performance, providing noise rather than a useful signal. Amongst the most important features are the aircraft's position, comm state, delta to the aircraft's XFL and direction. The least important features were the wake category, whether the aircraft was under ATC speed restrictions, the engine type, and the predicted time until sector exit. b) The green bars show the increase in MAE when first: scenario graphs are randomly connected instead of using our flight level overlap-based edge creation method; and second, when the scenario graphs contain no edges. The connectivity of the graph appears as consequential for the model prediction performance as the most important features. The error bars shown are 95\% confidence intervals calculated as 1.96 \texttimes\ the standard error of the mean.\vspace{-2mm}}\label{fig:feature_importance}
\end{figure}

Figure~\ref{fig:feature_importance} shows the results of the permutation analysis, described in Section \ref{sec:featureimportantceanalysis}, to determine the relevance of each node and edge feature. Figure~\ref{fig:feature_importance}a) shows the increase in MAE relative to the unperturbed dataset for each feature, ranked in order according to the performance degradation. A large positive change in the MAE (increase) compared to the original model is interpreted as the permuted feature having increased predictive power. Consequently, the most important features to the prediction model are the aircraft's lateral position, comm state, delta to XFL, lateral direction, current FL and sector exit direction. Amongst the least important features are the wake category, whether the aircraft was under ATC speed restrictions, the engine type, and the predicted time until sector exit. Interestingly, the edge features, separation distance and closing speed, have relatively low importance for the model. However, there is considerable redundancy in the model features. For example, the separation distance between two aircraft can still be inferred from the relative aircraft positions even with the specific distance feature scrambled. Equivalently, the closing speed can be gathered from the ground speed, position and lateral direction features.

Figure~\ref{fig:feature_importance}b) shows the results for randomly connected and edgeless graphs. The performance degradation when using random edges instead of our proposed graph construction method was greater than for any of the individual features. This shows that, despite the limited importance of edge features, the edges and graph connectivity structure itself were crucial to the predictive performance of the GNN model. Using scenario graphs with no edges at all also dramatically decreased the model performance further showing that the context of surrounding aircraft was crucial to the predictive accuracy of the trained GNN model.

\subsection{Quantitative baseline comparisons}
Table~\ref{tab:quatitative_comparison} contains the MAE results for the GNN model and the baseline approaches when evaluated on the reserved week of data from August 2019. The GNN model significantly outperformed the minimum-clearance baseline approach (Wilcoxon signed-rank test, $p < 10^{-12}$) in terms of the absolute error for the predicted number of clearances issued within the following 10-minute period. The GNN model slightly outperformed the random forest and XGBoost regression models in terms of MAE (Wilcoxon signed-rank test, $p = 0.04$ for the random forest and $p = 0.05$ for the XGBoost).
\begin{table}[ht]
\caption{\label{tab:quatitative_comparison} Mean absolute error (MAE) results for the different baseline methods. The GNN model significantly outperforms the minimum-clearance baseline method (Wilcoxon signed-rank test, $p < 10^{-12}$).}
\centering
\begin{tabular}{lc}
\rowcolor[RGB]{239,239,239}
\multicolumn{1}{c}{\textit{Method}} & \textit{MAE ($\pm 95$\% CI)} \\ \hline
minimum-clearance baseline  & $3.44 \pm 0.09$ \\
GNN model                   & $1.83 \pm 0.05$ \\
\rowcolor[RGB]{239,239,239}
\multicolumn{1}{c}{\textit{Regressor type}} &  \\ \hline
Random forest               & $1.87 \pm 0.04$ \\
XGBoost                     & $1.86 \pm 0.04$ \\
\end{tabular}
\end{table}

\begin{figure}[h!]
\centering
\includegraphics[width=\textwidth]{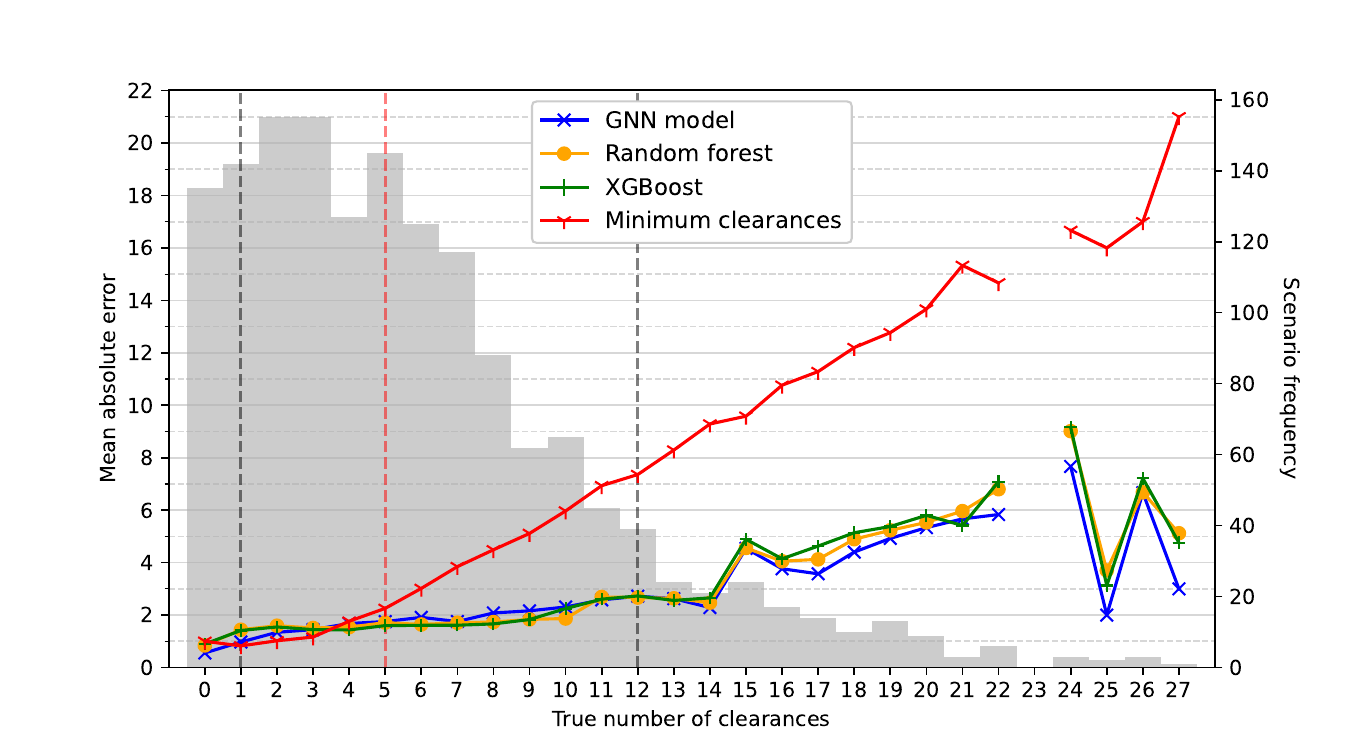}
\caption{MAE values separated by ground truth clearance value for the GNN, random forest, XGBoost and minimum-clearance baseline approaches. The underlying distribution of number of scenarios containing each of these number of issued clearances is shown with the grey histogram. The dashed lines indicate the 10\textsuperscript{th} percentile, median and 90\textsuperscript{th} percentile values for this target distribution.\vspace{-2mm}}\label{fig:distribution_preds}
\end{figure}

Figure~\ref{fig:distribution_preds} shows the MAE results for the GNN method, random forest and XGBoost regressor approaches and the minimum-clearance baseline over the range of the target distribution of number of clearances issued. Behind the lines indicating the MAE for each method we show a histogram of the number of scenarios with each of number clearances. Dashed lines have been included to show the 10\textsuperscript{th} percentile, median and 90\textsuperscript{th} percentile of the target distribution, and we will discuss the results in terms of the ranges between these indicated quantiles. In terms of numbers of issued clearances, these regions correspond to 0 to 1 clearances, 2 to 5 clearances, 6 to 11 clearances and $\geq$12 clearances issued in the ten-minute period following the input scenario.

All four methods perform similarly in the lower end of the distribution where very few clearances are issued. Such scenarios are likely to be easy to identify as the ATCO was not required to take many actions. Above the median value of three clearances issued, the accuracy of the minimum-clearance baseline worsens substantially, evidence that the scenarios in this part of the target distribution require more input from the ATCO than simply clearing each aircraft to the required sector exit flight level. This suggests a greater number of aircraft interactions and complexity that will increase the task demand for the ATCO in this range. The GNN performance is slightly worse than the random forest and XGBoost methods in the range between the median and 90\textsuperscript{th} percentile value (6 - 11 clearances issued) but is superior to both regression methods in the long tail of the distribution. In the region beyond the 90\textsuperscript{th} percentile ($\geq$12 clearances issued), the random forest and XGBoost median signed error was -3.9 and -3.6, showing systematic under-estimation of the load compared to -2.0 for the GNN method. We suggest this is because of our weighted sampling strategy to expose the GNN model to more scenarios with higher clearance counts at training time. It is these scenarios in which a high number of clearances were issued that we anticipate ATCO was undergoing the most task demand, and therefore the scenarios it was most impactful to correctly identify.

One of the key assumptions of this work is that the number of issued clearances can be used as a proxy measure for ATCO task demand. This assumption is less strong for scenarios with a lower number of issued clearances, where it is more difficult to discern the true task demand from the number of clearances alone. For example, the cognitive load on an ATCO of issuing three or four clearances within the ten-minute period could be low or high, as they could correspond to fairly routine operations or a particularly intricate and well-designed set of clearances to solve a complex set of aircraft interactions. The assumption becomes stronger as the number of issued clearances increases. For example, if the aircraft in the scenario require 15 or more clearances over a 10-minute period, this is more likely to be cognitively demanding for the ATCO. It is expected that scenarios requiring such elevated levels of control will contain more complex aircraft interactions and increased ATCO task demand.

\FloatBarrier
\subsection{Qualitative comparison to alternative task demand prediction tools}
\begin{figure}[ht]
\centering
\includegraphics[width=\textwidth]{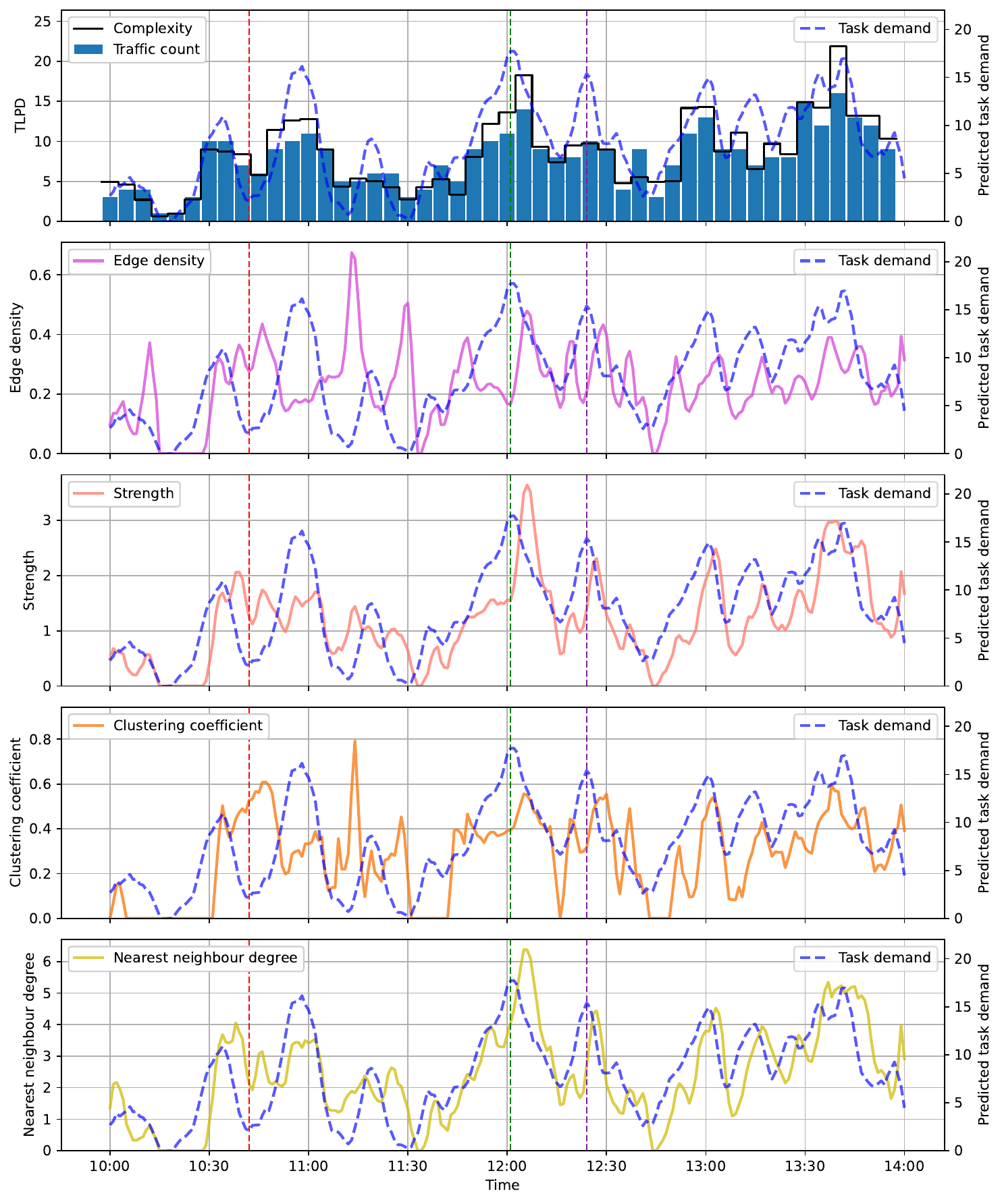}
\caption{Longitudinal comparison over a period of four hours of the established methods for complexity prediction, TLPD and graph indicators, with our GNN-based ATCO task demand estimations. The top row shows the TLPD traffic count in the histogram bars and complexity prediction with the black line. Our estimations of task demand are shown with the dashed blue line. Each subsequent row shows one of the four graph indicators, \textit{edge density}, \textit{strength}, \textit{clustering coefficient} and \textit{nearest neighbour degree} and the estimated task demand. The red, green and purple dashed lines indicate the times of the example scenarios shown in Fig.~\ref{fig:qualitative_results_examples}.}\label{fig:qualitative_results}
\end{figure}

While the quantitative results validate our model's predictive accuracy, the limitations of using clearance counts as a proxy for complexity motivate a deeper, qualitative analysis. We now evaluate the core contribution of our framework: its ability to produce an interpretable, per-aircraft measure of task demand using the ablation method described in Sec.~\ref{sssec:taskdemandestimation}. To do this, we compare our task demand estimations against the established TLPD tool and the graph-based indicators from \citeauthor{Isufaj2021} over a four-hour operational period. Figure~\ref{fig:qualitative_results} shows this comparison.

The time period between 10:00 and 14:00 on a day from the reserved test set of data is shown because it contained periods of both high and low activity within LMS. The GNN model-estimated task demand shown is $\sum\phi_i$ computed at one-minute intervals (using Eq.~\eqref{eq:taskdemandeq}), i.e. the sum of the number of clearances each aircraft contributes in the scenario. As mentioned previously, the TLPD values have two components: the traffic count, which is the number of aircraft present in the sector within a five-minute window, and a predicted complexity. The height of the complexity line relative to the traffic count indicates the estimated complexity of the scenario for that amount of traffic. The edge density, strength, clustering coefficient, and nearest neighbour degree graph indicators of \citeauthor{Isufaj2021} are shown on consecutive plots with the predicted task demand for comparison.

\begin{figure}[ht]
\centering
\vspace{-1.5mm}
\includegraphics[width=\textwidth]{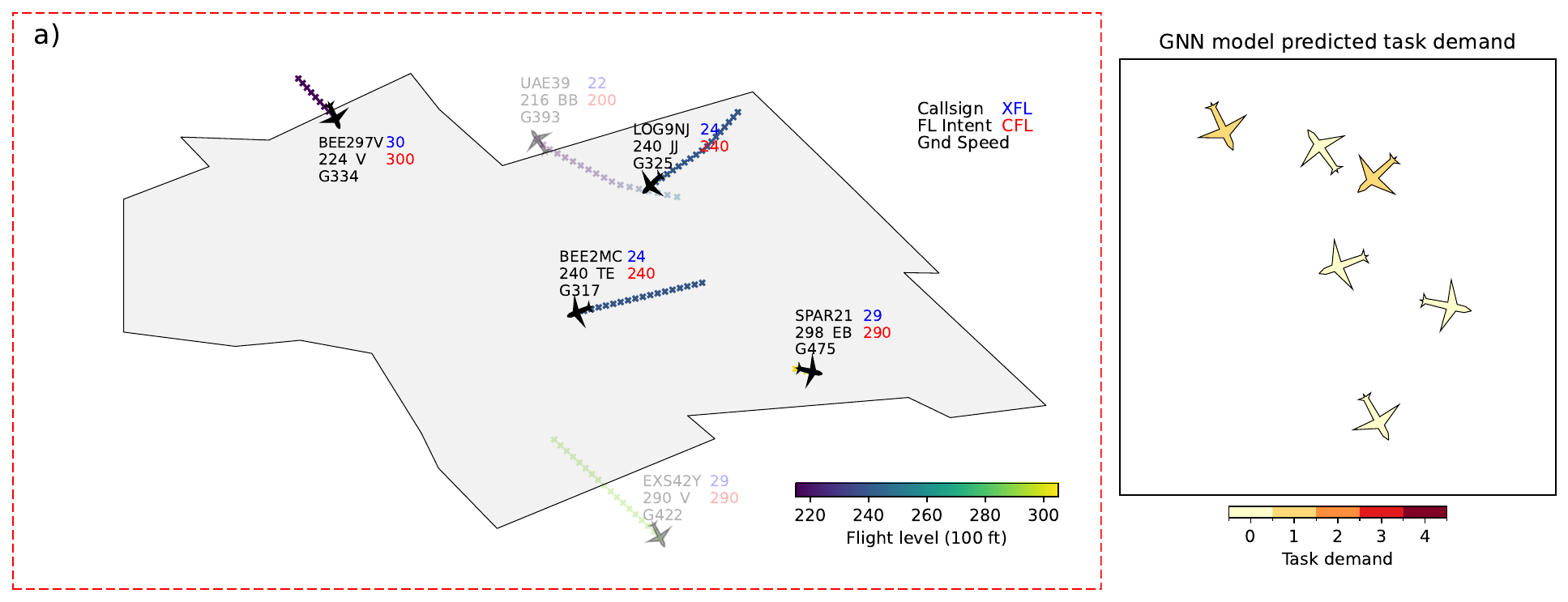}
\vspace{-1.5mm}
\includegraphics[width=\textwidth]{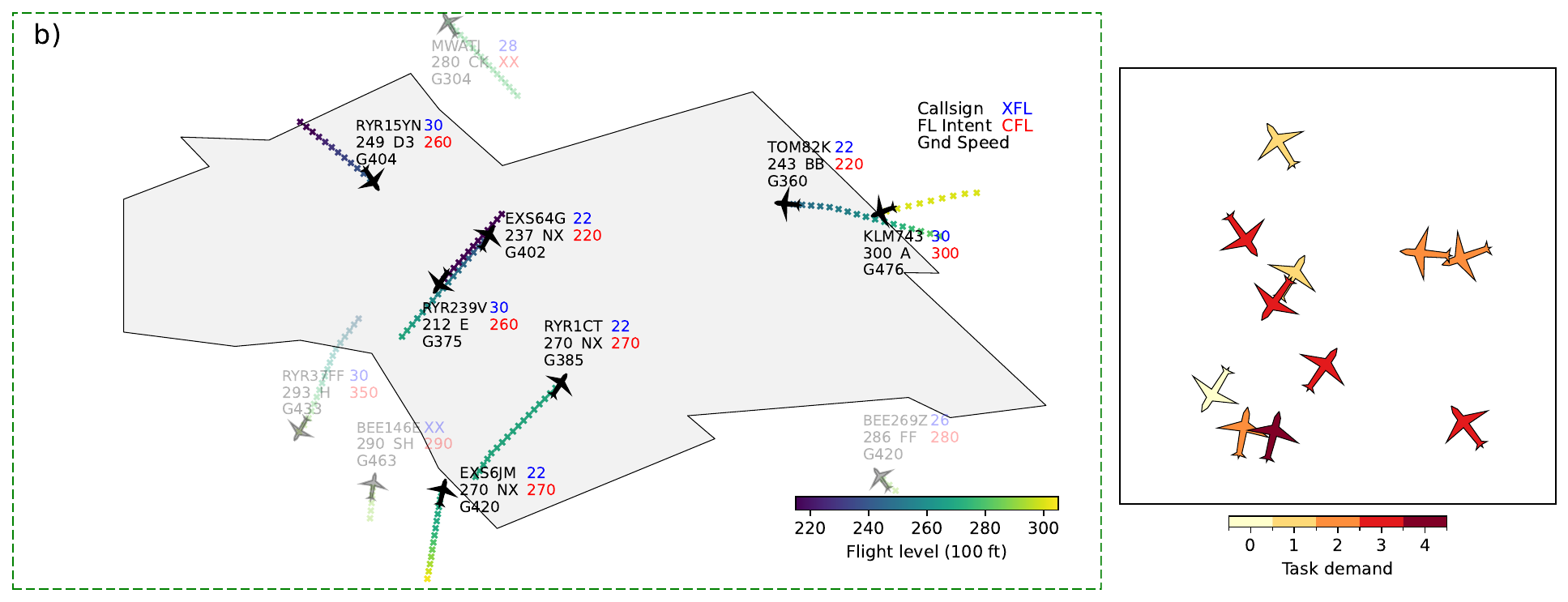}
\vspace{-1.5mm}
\includegraphics[width=\textwidth]{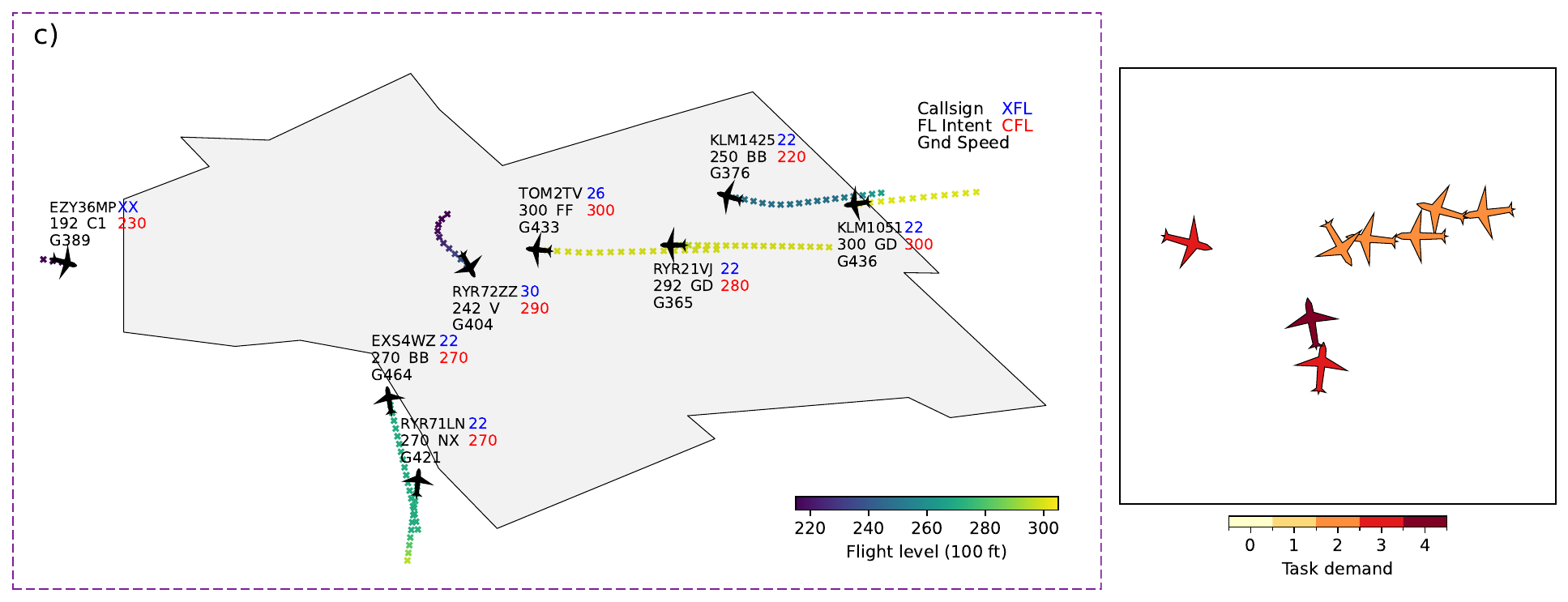}
\vspace{-4mm}
\caption{Three example scenarios corresponding to the times indicated by the red, green and purple dashed lines in Figure~\ref{fig:qualitative_results}. Sector maps with the present aircraft are shown on the left, with per-aircraft estimations of the ATCO task demand from our GNN-based method shown on the right.
Example a) shows a fairly quiet scenario, where all aircraft have been cleared to their intended exit flight level. Accordingly, the predicted task demand is relatively low for all aircraft, excluding aircraft BEE297V which needs to cross the whole sector. Example b) shows the sector with a larger amount of traffic, with several climbing and descending aircraft crossing the centre of the sector. This scenario will likely require closer ATCO attention and the estimated task demand is elevated in agreement. Example c) shows another time at which several aircraft are climbing and descending through the centre of the sector. In this particular example, the ATCO will have had to very carefully sequence multiple aircraft headed north around aircraft headed west. The predicted task demand at this time is also high suggesting a fairly complex air traffic scenario.}\label{fig:qualitative_results_examples}
\end{figure}

The predicted task demand and TLPD outputs visually align fairly well over the presented time period. Many of the peaks in TLPD complexity, when the black line is above the traffic count histogram, are accompanied by a predicted increase in task demand e.g. 10:55, 12:00, and 13:40. However, there are instances of discrepancies between the two methods, such as 12:25 where the task demand is predicted to be high but the TLPD model suggests a scenario of average complexity (see example ~\ref{fig:qualitative_results_examples}c)).

Across the four-hour span there is a fairly consistent lag between peaks in GNN-predicted task demand and peaks in the graph indicator measures. The task demand represents a short-term forecast of anticipated complexity, based on a prediction of the number of clearances required in the upcoming ten-minute period. In contrast, the graph indicators are metrics reflecting the instantaneous state of a scenario. The indicator values cannot be easily shifted earlier to serve as predictive values, as this would require advanced trajectory prediction methods which would introduce additional uncertainties.
The lag observed between the task demand and graph indicator signals likely reflects this fundamental difference between a predictive model and a real-time measurement.
To quantify the observed lag time, the Pearson correlation coefficient between the GNN task demand and each of the graph indicators was calculated at 1-minute interval offsets between 0 and 20 minutes. The correlation coefficients were maximised when the indicator output values were shifted forward by between 3-5 minutes.

There are some noticeable similarities in the patterns of task demand predictions and graph indicators over the four-hour span. Despite this, no single indicator aligns consistently with the task demand. \citeauthor{Isufaj2021} argue that the four indicators together present a rounded picture of the airspace complexity. However, the abstract nature of the graph indicators makes it difficult to draw direct conclusions about scenario complexity. In contrast, our predicted task demand offers a more intuitive and operationally relevant measure.

The red, green, and purple vertical dashed lines in Fig.~\ref{fig:qualitative_results} show the times at which three example snapshots have been extracted. These scenario examples are shown in Fig.~\ref{fig:qualitative_results_examples}. Each example has the sector map showing the aircraft present at that time and the per-aircraft task demand estimated by the GNN-based approach.

Example ~\ref{fig:qualitative_results_examples}a), outlined in red, is taken from around 10:40. This example was selected because both TLPD and our task demand predictions are in agreement and are both relatively low. In contrast, the graph indicators are relatively high with the clustering coefficient indicators reaching a peak value at this point. As can be seen in the figure, the scenario has a below-average level of traffic, with only four aircraft in communication with the ATCO and two that have already switched frequency and are in communication with the next sector on their route (UAE39 and EXS42Y). The ATCO issued only five clearances to these aircraft over the next ten-minute period, further confirming that this example is relatively simple and corroborating the outputs of TLPD and our task demand prediction. The full computation graph for the indicators is shown in the appendix for this example snapshot to demonstrate why particular graph indicators are high at this time.

In example ~\ref{fig:qualitative_results_examples}b), outlined in green, just after 12:00, the predicted task demand is high, as is the relative complexity according to the TLPD outputs. All four of the graph indicators also have peaks, but five minutes after the peak predicted by our approach. At this time there are seven aircraft in the sector in communication with the ATCO, and a further two about to enter (BEE146E and BEE269Z). In this scenario, 26 clearances were issued to these aircraft over the next ten-minute period, indicating an elevated level of complexity compared to the first example. The aircraft with the highest task demand predictions from the GNN approach recently entered the sector (EXS6JM and BEE269Z) or were heading towards the middle of the sector with overlapping flight levels (RYR1CT, RYR15YN, TOM82K). Aircraft MWATJ and RYR37FF had recently been passed to the neighbouring sector and duly require no further clearances by the ATCO.

Example ~\ref{fig:qualitative_results_examples}c) shows the final highlighted scenario at around 12:25. At this time, the predicted task demand from our GNN-based approach reaches a peak while the TLPD measure suggests the scenario complexity is about average. The four graph indicators have similar peaks, but again with lag times ranging between $2$ and $5$ minutes. Similarly to the previous example, the aircraft predicted to contribute most highly to the ATCO task demand are those that have recently entered the sector (EZY36MP, KLM1051, EXS4WZ and RYR71LN). Each of these aircraft requires climbing or descending whilst being routed through the middle portion of the sector. The RYR21VJ and KLM1051 aircraft in the east of the sector are destined for Bristol airport and, therefore, will transit and exit the sector to the west. The EXS4WZ and RYR71LN aircraft to the south will be exiting at the north of the sector because they are travelling to Birmingham and East Midlands airports respectively. This intricate traffic movement will have required careful coordination by the ATCO, who issued $18$ clearances to aircraft in this scenario over the following $10$-minute period.

\vspace{2mm}

Table~\ref{tab:correlation_coefficients} contains the Pearson correlation coefficients between the task demand estimation methods, the number of clearances issued and the traffic count. The first column shows that the four graph indicator measures and the existing TLPD complexity metric are sufficiently different to our proposed GNN-based task demand predictor. In the second and third columns we compare how the approaches correlate with the number of clearances issued following the input scenario and with the traffic count, i.e. the number of aircraft present in the sector. Each of the approaches correlates more strongly with the traffic count than the number of clearances issued (shown in red), apart from our GNN-based method (shown in green). This finding suggests that the TLPD complexity and graph indicators are heavily influenced by raw traffic count, whereas our GNN-based method is more sensitive to the underlying interactive complexity of a scenario. Our approach therefore appears to be a more reliable estimator of ATCO task demand.

The Pearson correlation values between the graph indicators are shown in full in the appendix. They are largely consistent with the original analysis of \citeauthor{Isufaj2021}, suggesting we employed a faithful re-implementation of the indicators~\citep{Isufaj2021}.

\begin{table}[h]
\caption{\label{tab:correlation_coefficients} Pearson correlation coefficients between GNN-based model task demand estimations, TLPD traffic count, complexity predictions, and graph indicators.}
\centering
\renewcommand{\arraystretch}{1}
\setlength{\tabcolsep}{4pt} 
\small 
\begin{tabular}{cc|cc}
\toprule
 & GNN task demand & $N_{\text{clearances}}$ & TLPD traffic count \\
\hline
GNN task demand & - & \cellcolor[RGB]{179,242,179}\textbf{0.91} & 0.74 \\
\hline
TLPD complexity & 0.79 & 0.76 & \cellcolor[RGB]{255,179,179}\textbf{0.94} \\
ED & 0.18 & 0.04 & \cellcolor[RGB]{255,179,179}\textbf{0.42} \\
Strength& 0.67 & 0.51 & \cellcolor[RGB]{255,179,179}\textbf{0.80} \\
CC& 0.50 & 0.31 & \cellcolor[RGB]{255,179,179}\textbf{0.61} \\
NND& 0.72 & 0.56 & \cellcolor[RGB]{255,179,179}\textbf{0.82} \\
\bottomrule
\end{tabular}
\end{table}

\newpage

\newpage
\FloatBarrier

\section{Conclusion}
In this paper, we introduced a GNN-based framework to estimate future ATCO task demand. By leveraging graph representations of air traffic scenarios, our approach effectively captured relational dependencies between aircraft, enabling better-informed and context-aware predictions. We validated the model's performance by comparing its ability to predict short-term upcoming clearances against an ATCO-informed heuristic and two regression baselines, achieving superior predictive accuracy. These results validate the use of graph structures to explicitly model aircraft interactions. Moreover, through specific air traffic scenarios, we demonstrated that our method offers an \textit{interpretable} alternative to traditional complexity indicators. Importantly, while our framework can readily incorporate domain-driven features, it does not require hand-crafted factors such as known hotspots to achieve strong predictive performance.

Looking ahead, we propose developing more quantitative metrics to evaluate the success and practical relevance of our task demand measure. In particular, we envision collecting a labelled dataset where ATCOs explicitly identify which aircraft they consider relevant in a given scenario, especially in relation to a specific aircraft. Such data would enable a direct assessment of the GNN's ability to highlight high-priority aircraft and quantify alignment with expert human judgement. Future efforts should focus on evaluating the method's generalisability by applying it to other airspace sectors with different traffic patterns and control strategies. Additionally, future work could explore how best to incorporate the extra fidelity of our approach, for example, investigating the \textit{weights} of predicted interactions between aircraft that are relevant to the ATCOs task. Leveraging datasets that include scheduled arrival times could also extend the prediction horizon, allowing the model to make complexity estimates further into the future.

In summary, while predictive performance remains an important benchmark, our findings suggest that the true strength of attention-based graph modelling in air traffic management lies in its interpretability and its power to uncover interaction-driven notions of complexity and priority. Moving forward, we see significant value in combining predictive modelling with targeted ablation studies to further validate and refine the proposed approach.

\section*{Appendix}
\subsection*{Cross Validation}
\label{sec:xvalidation}
Table~\ref{tab:MAE_fivefold} shows the results of the five-fold cross-validation for models trained with all the features listed in Table~\ref{tab:features}. Each model was assessed using the test set specific to that fold, and we show the mean absolute error between the predicted number of clearances and the actual number issued. The best performing model (fold 2) was used to evaluate the relative importance of the input features, as it was only computationally feasible to run permutation tests for 18 features (16 node features, 2 edge features) on 125{\small,}020 scenarios for a single model. However, this was considered acceptable as we observed fairly consistent cross-validation performance across each of the five models.

\begin{table}[ht]
\caption{Mean absolute error (MAE) for each model trained in the five-fold cross-validation within the test set for that fold. The model trained in fold two achieved the lowest MAE and was used to evaluate input feature importance. The values reported are the mean $\pm$ 95\% confidence intervals (CI), calculated as 1.96 \texttimes\ standard error of the mean.}
\label{tab:MAE_fivefold}
\centering
\begin{tabular}{cccccc}
\rowcolor[RGB]{239,239,239}
\textit{Cross-validation fold number} & \textit{1} & \textit{2} & \textit{3} & \textit{4} & \textit{5} \\ \hline
MAE ($\pm 95$\% CI) & $1.79\pm0.05$ & $1.77\pm0.05$ & $1.83\pm0.05$ & $1.82\pm0.04$ & $1.77\pm0.05$
\end{tabular}
\end{table}

\subsection*{Graph Indicators supplement}
\label{sec:indicators_example}

\begin{figure}[t]
\centering
\includegraphics[width=\textwidth]{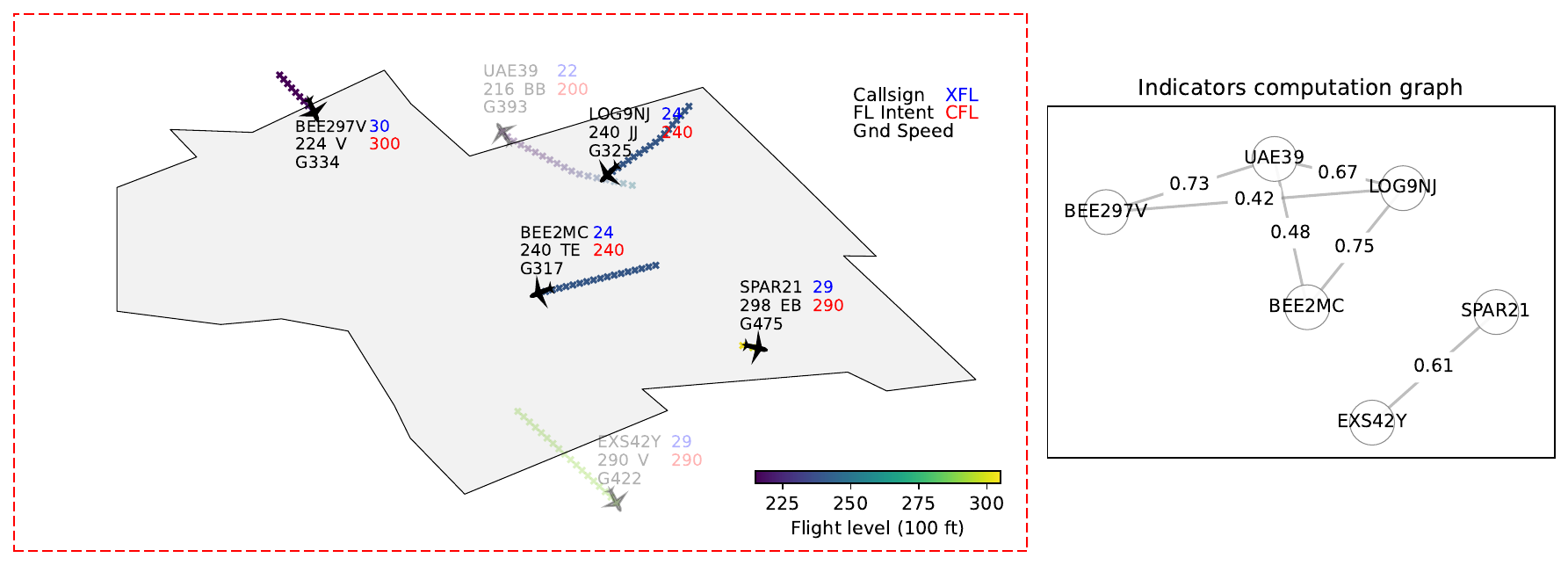}
\caption{The example scenario corresponding to the time indicated by the red dashed lines in Fig.~\ref{fig:qualitative_results}. This time the indicators computation graph is included on the right-hand side to show the edges and weights between aircraft. Each node in the computation graph corresponds to an aircraft in the scenario and edges are created following the process outlined in Sec.~\ref{sssec:quali_baselines}. Some aircraft nodes have been moved slightly in the visual representation for clarity.}\label{fig:appendix_indicators_example}
\end{figure}

Figure~\ref{fig:appendix_indicators_example} shows the computation graph for the indicators constructed by \citet{Isufaj2021} at the time of the first example scenario shown in Fig.~\ref{fig:qualitative_results_examples}. For this particular scenario, all the graph indicators are relatively high, with the clustering coefficient in particular reaching peak values. This is despite a relatively low traffic count, reduced complexity predicted by the TLPD and low predicted task demand from the GNN-based approach. The indicators' computation graph is relatively well-connected, with edges between most aircraft in the sector, and all with weights $>0.4$, which accounts for the increased edge density indicator. The presence of two tightly clustered triplets, containing highly weighted edges, explains the high clustering coefficient indicator for this scenario.

\newpage

The Pearson correlation values between the graph indicators are shown in Table~\ref{tab:graph_indicator_correlation_coefficients} and are broadly consistent with the original analysis of \citet{Isufaj2021}.

\begin{table}[ht]
\caption{\label{tab:graph_indicator_correlation_coefficients} Pearson correlation coefficients between each of the graph indicators.}
\centering
\renewcommand{\arraystretch}{1}
\setlength{\tabcolsep}{4pt} 
\small 
\begin{tabular}{ccccc}
\toprule
 & ED & Strength & CC & NND \\
\midrule
ED          &      & 0.68 & 0.79 & 0.61 \\
Strength    & 0.68 &      & 0.86 & 0.98 \\
CC          & 0.79 & 0.86 &      & 0.82 \\
NND         & 0.61 & 0.98 & 0.82 &      \\
\bottomrule
\end{tabular}
\end{table}

\FloatBarrier

\section*{Acknowledgments}
This work is supported by the grant “EP/V056522/1: Advancing Probabilistic Machine Learning to Deliver Safer, More Efficient and Predictable Air Traffic Control” (aka Project Bluebird), an EPSRC Prosperity Partnership between NATS, The Alan Turing Institute, and the University of Exeter. Generative AI tools were used sparingly in the preparation of this manuscript to improve grammar and clarity.

\bibliography{abstract}
\end{document}